\documentclass[journal]{IEEEtai}
\usepackage[colorlinks,urlcolor=blue,linkcolor=blue,citecolor=blue]{hyperref}

\usepackage{color,array}

\usepackage{amssymb}
\usepackage{lipsum}
\usepackage{multirow}
\usepackage{amsmath}
\usepackage{graphicx}
\usepackage{float} 
\usepackage{subcaption}
\usepackage{caption}
\let\labelindent\relax
\usepackage{enumitem}
\usepackage{hyperref}
\usepackage{booktabs}
\usepackage[table,xcdraw]{xcolor}

\begin{document}

\title{Video-Based MPAA Rating Prediction: An Attention-Driven Hybrid Architecture Using Contrastive Learning} 


\author{Dipta Neogi, Nourash Azmine Chowdhury, Muhammad Rafsan Kabir, Mohammad Ashrafuzzaman Khan 
\thanks{Dipta Neogi, Nourash Azmine Chowdhury, Muhammad Rafsan Kabir, and Mohammad Ashrafuzzaman Khan are with the Department of Electrical and Computer Engineering at North South University, Dhaka, 1229, Bangladesh. \emph{Corresponding author: Muhammad Rafsan Kabir} (E-mail: muhammad.kabir@northsouth.edu)}}


\maketitle

\begin{abstract}
The rapid growth of visual content consumption across platforms necessitates automated video classification for age-suitability standards like the MPAA rating system (G, PG, PG-13, R). Traditional methods struggle with large labeled data requirements, poor generalization, and inefficient feature learning. To address these challenges, we employ contrastive learning for improved discrimination and adaptability, exploring three frameworks—Instance Discrimination, Contextual Contrastive Learning, and Multi-View Contrastive Learning. Our hybrid architecture integrates an LRCN (CNN+LSTM) backbone with a Bahdanau attention mechanism, achieving state-of-the-art performance in the Contextual Contrastive Learning framework, with 88\% accuracy and an F1 score of 0.8815. By combining CNNs for spatial features, LSTMs for temporal modeling, and attention mechanisms for dynamic frame prioritization, the model excels in fine-grained borderline distinctions, such as differentiating PG-13 and R-rated content. We evaluate the model's performance across various contrastive loss functions, including NT-Xent, NT-logistic, and Margin Triplet, demonstrating the robustness of our proposed architecture. To ensure practical application, the model is deployed as a web application for real-time MPAA rating classification, offering an efficient solution for automated content compliance across streaming platforms.
\end{abstract}

\begin{IEEEImpStatement}
With the exponential rise in digital content, ensuring age-appropriate media consumption has become a global challenge. Our research introduces a novel AI-driven framework for automated MPAA rating classification of videos, addressing the inefficiencies of manual content review. By leveraging a hybrid deep learning model—integrating CNN, LSTM, and attention mechanisms within a contrastive learning paradigm—our approach achieves state-of-the-art accuracy (88\%) while maintaining computational efficiency. This proposed method offers benefits across multiple dimensions: \emph{(a)} Socially, it enhances online safety by filtering content for young viewers; \emph{(b)} Technologically, it advances AI video analysis, with applications extending beyond ratings; \emph{(c)} Environmentally, its computationally efficient architecture promotes sustainable computing.

\end{IEEEImpStatement}


\section{Introduction}

\IEEEPARstart{A}{s} digital platforms like Netflix, Disney+, and YouTube expand \cite{wayne2018netflix}, ensuring that younger audiences access only age-appropriate content has become increasingly critical \cite{anghelcev2021binge}. The MPAA rating system \cite{county95motion}, with classifications such as G, PG, PG-13, and R, helps families make informed viewing choices \cite{septimus1996mpaa}, protecting young viewers from exposure to inappropriate content \cite{falkowski2020current}. However, the sheer volume of daily uploads \cite{arriagada2020you} makes manual rating methods impractical. Streaming platforms require automated classification systems to ensure compliance with MPAA guidelines and global regulations while maintaining consistency and accuracy. These systems help prevent regulatory issues and public backlash, ultimately enhancing viewer safety and experience.

Recent advancements in machine learning and deep learning have enabled automated systems to predict MPAA ratings with remarkable accuracy. Masha et al. \cite{shafaei2019rating} proposed multimodal approaches integrating textual data from movie trailers, while Ha et al. \cite{mohamed2021combining} combined film scripts with static visual features, highlighting the benefits of diverse data modalities. However, these studies primarily focus on textual and static visual data, overlooking the complexity of dynamic video representations, which inherently capture temporal and contextual cues. Uzzaman et al. \cite{uzzaman2022lrcn} and Uddin et al. \cite{uddin2024deep} developed ConvLSTM and LRCN-based models for human activity recognition, achieving high accuracy on UCF50 and HMDB51 datasets. Despite progress in video classification, these approaches neglect factors like environmental noise and video quality, which impact real-world performance. This underscores the need for more efficient, scalable solutions adaptable to diverse and dynamic environments.

To address the aforementioned gaps, we first curated a custom video dataset featuring 323 diverse samples across four MPAA rating classes (G, PG, PG-13, and R). We trained and evaluated existing video classification models, such as ResNet3D-50 \cite{tran2018closer} and LRCN (LSTM + CNN) \cite{pandya2021segregating} in contrastive learning framework, on the curated dataset for classifying MPAA ratings. Then, for further robust video classification, we incorporated an additional attention mechanism to the existing LRCN backbone, introducing a novel hybrid backbone in contrastive learning for video classification. This architecture is designed to capture both the visual details (spatial features) and the flow of the video over time (temporal features). The attention mechanism helps the model focus on the most important frames in a sequence, enabling it to detect subtle differences, such as those distinguishing a PG-13 video from an R-rated one. In our study, we employ three distinct attention mechanisms: self-attention \cite{vaswani2017attention}, co-attention \cite{lu2016hierarchical}, and Bahdanau attention \cite{bahdanau2014neural}. This framework allows the model to learn robust, context-aware, and finely tuned video representations. To ensure accurate differentiation between MPAA rating classes, various methods are applied during the training process to minimize confusion, particularly for videos that may fall on the boundary between two categories. The model is designed to be both efficient and lightweight, making it highly suitable for use on memory-constrained devices without compromising performance. Moreover, we developed a web application to demonstrate the real-world usability of our proposed model architecture for MPAA rating classification from videos. 

\noindent \textbf{Contributions:} The key contributions are as follows:

\begin{itemize}
    \item We curated a custom video dataset from diverse internet sources, comprising 323 video samples across four MPAA rating classes (G, PG, PG-13, R), with durations ranging from 11 to 25 seconds in various formats to ensure heterogeneity and robustness.
    \item A novel hybrid model architecture is proposed within a contrastive learning framework that incorporates an LRCN backbone, consisting of CNN and LSTM, with an attention mechanism to improve adaptability, temporal coherence, and representation granularity for video classification.
    \item Advanced attention mechanisms—self-attention, co-attention, and Bahdanau attention—are integrated to selectively emphasize contextually significant features, enabling interpretable and context-aware embeddings for managing complex cross-frame dependencies.
    \item Several contrastive loss functions, including NT-logistic, NT-Xent, and Margin Triplet, are employed to optimize embedding quality during pre-training and fine-tuning, with gradient clipping ensuring stability and convergence during training.

    \item We evaluated the performance of our proposed model architecture along with existing architectures within the contrastive learning framework by analyzing their execution time, model size, number of parameters, and overall classification performance.
    
\end{itemize}

\section{Related Works}

\begin{figure*}[!t]
    \centering
    \includegraphics[width=0.8\textwidth]{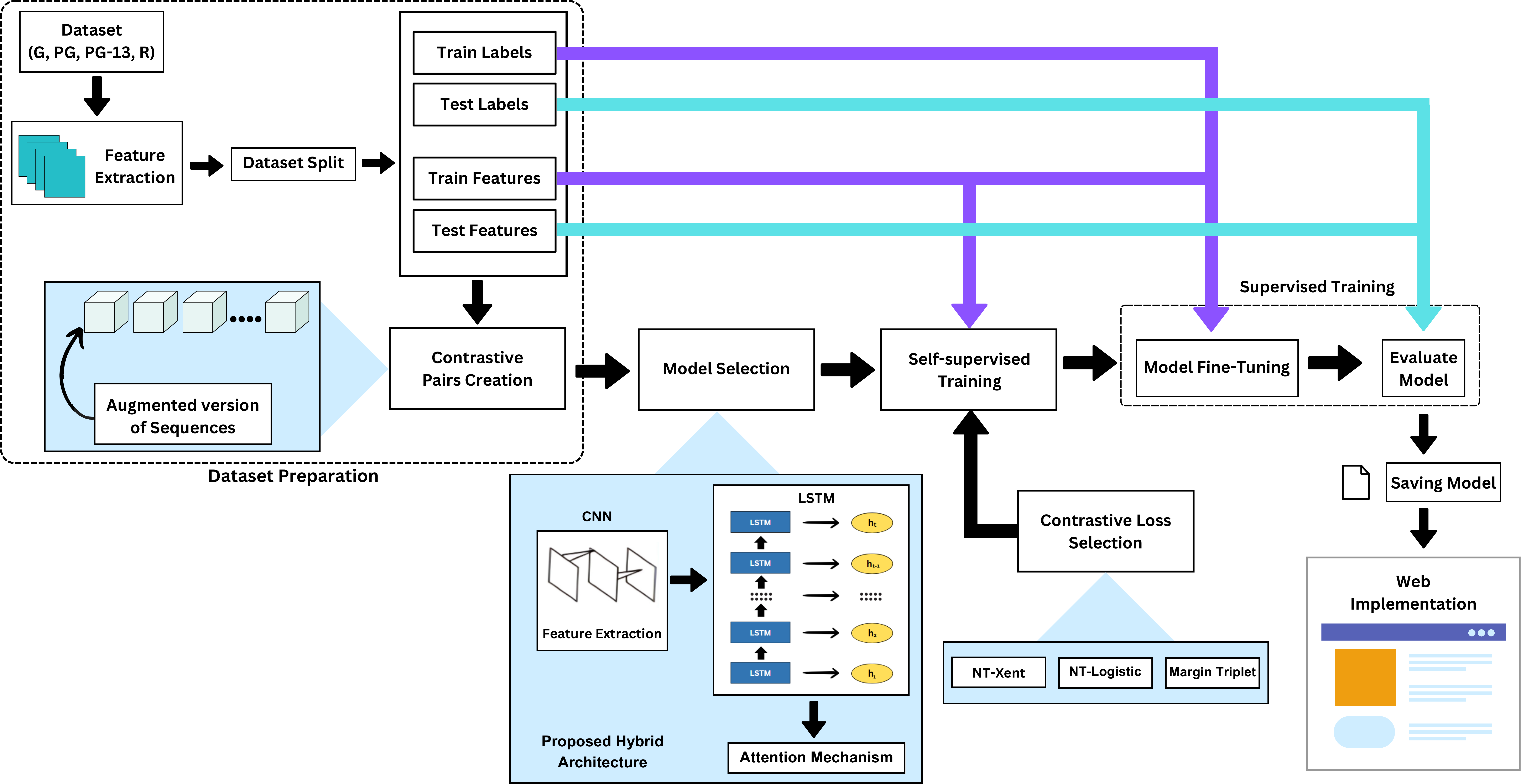}
    \caption{ Overall Pipeline for Contrastive Learning with Hybrid Architectures. This illustration presents the comprehensive pipeline for contrastive learning, from dataset preparation and feature extraction using CNNs through model pre-training and fine-tuning with LSTM and attention mechanisms. It includes steps like contrastive pair creation, model evaluation, and web implementation, illustrating the end-to-end process from data to deployment}
    \label{fig:workflow}
\end{figure*}

\noindent\textbf{Multimodal and metadata-based applications:} Recent video classification research has increasingly focused on multimodal approaches that combine visual content with metadata or textual information. Several studies exemplify this trend: Masha et al. \cite{shafaei2021case} introduced a movie trailer dataset with age ratings, using multimodal learning for automated suitability assessments. Similarly, Ha et al. \cite{mohamed2021combining} developed a bi-modal model integrating film scripts and IMDb image features to predict age ratings, demonstrating the value of combining textual and visual data. Shafaei et al. \cite{shafaei2019rating} further advanced this direction by employing RNNs with attention mechanisms on scripts to predict MPAA ratings, achieving a 78\% F1 score through genre and emotion analysis. However, these approaches share critical limitations: they depend heavily on metadata availability, require complex multimodal fusion pipelines, and may not generalize well to video-only datasets. Our work addresses these limitations by eliminating metadata dependencies entirely, focusing instead on self-supervised contrastive learning from raw video frames. This makes our approach more versatile for scenarios where only visual content is available.

\noindent\textbf{Self-supervised and contrastive learning in video representation:} Self-supervised learning has emerged as a powerful paradigm for learning video representations without manual annotations. Recent advances include Qian et al.'s CVRL \cite{qian2021spatiotemporal}, which uses spatial-temporal augmentations with 3D-ResNet-50 to achieve 70.4\% top-1 accuracy on Kinetics-600 \cite{kay2017kinetics}, outperforming ImageNet-supervised methods. Moreover, Diba et al. \cite{diba2021vi2clr} further improved self-supervised video representation through Vi²CLR, combining ResNet-50 and S3D ConvNet encoders with clustering-based supervision to surpass MoCo \cite{he2020momentum} and SimCLR \cite{chen2020simple}. Similarly, Yang et al. \cite{yang2020video} introduced VTHCL, leveraging visual tempo for self-supervision to achieve 82.1\% on UCF-101 \cite{soomro2012ucf101}, while Hernandez et al. \cite{hernandez2025vic} combined MAE with contrastive learning (ViC-MAE) to boost Kinetics-400 \cite{kay2017kinetics} performance to 81.50\%. Despite their successes, these methods face challenges including potential overfitting with large backbones, sensitivity to dataset scale, and reliance on predefined augmentation strategies. Our approach overcomes these limitations by incorporating attention mechanisms and a novel multi-view training framework that better captures temporal dependencies without requiring handcrafted augmentations or masking strategies.

\noindent\textbf{Hybrid architectures and temporal modeling:} The integration of hybrid architectures has become a prominent strategy for improving temporal modeling in video understanding. Notable examples include Jenni et al. \cite{jenni2023audio}, who combined audio-visual contrastive learning to achieve state-of-the-art performance on UCF101 \cite{soomro2012ucf101} and HMDB51 \cite{kuehne2011hmdb}. In addition, Zhu et al. \cite{zhu2024efficient} proposed a multimodal transformer (MMT) with novel contrastive losses, demonstrating significant accuracy improvements through audio-video fusion. Additionally, Wang et al. \cite{wang2023molo} developed MoLo for few-shot action recognition, employing a long-short contrastive objective with motion auto-decoding to enhance frame features. While effective, these hybrid approaches incur substantial computational overhead from multimodal fusion and face challenges in abstract action recognition. Our method simplifies the architecture by focusing exclusively on video data, combining the efficiency of LRCN (CNN + LSTM) with attention mechanisms in a contrastive learning framework. This specialized design enables more efficient temporal feature extraction while maintaining strong performance on pure video tasks.

\section{Methodology}
This section details the dataset used and outlines our proposed hybrid model architecture, which integrates Convolutional Neural Networks (CNNs), Long Short-Term Memory (LSTM) networks, and attention mechanisms. Additionally, it describes our approach to contrastive learning, covering the generation of contrastive pairs, the calculation of contrastive loss, and the model training process. The overall contrastive learning pipeline, illustrated in Figure \ref{fig:workflow}, presents a comprehensive workflow—from dataset preparation and CNN-based feature extraction to LSTM and attention-driven model refinement—highlighting key steps from contrastive pair creation to deployment.

\subsection{Dataset}
To develop an effective system for MPAA rating classification from video clips, we construct a custom dataset by collecting video and movie clips from various online sources. The dataset consists of four categorical classes corresponding to the MPAA rating system: G (suitable for all audiences), PG (parental guidance suggested), PG-13 (parents strongly cautioned for audiences under 13), and R (restricted to mature audiences). Each class contains a balanced number of samples, with 82, 80, 80, and 81 video clips, respectively. The duration of the video clips in the dataset ranges from 11 to 25 seconds, offering a compact temporal window for analysis while preserving essential contextual information for classification. To ensure diversity and robustness in data representation, the clips are sourced from a wide array of internet repositories and include multiple file formats such as avi, mp4, and other commonly used extensions. This heterogeneity in video formats provides a comprehensive basis for evaluating the model's performance across varying data types and structures. By curating a dataset that encompasses both a wide range of MPAA classes and a variety of content formats, we aim to provide a reliable benchmark for evaluating automated MPAA rating classification systems.

To enhance the diversity of the input data, we employed data augmentation techniques by applying some transformations. These augmentation techniques improve the model's robustness by making it invariant to changes in orientation, lighting, and contrast. The specific details of the augmentation process are provided in Table \ref{tab:Augment}.

\begin{table}[!t]
\small
\centering
\caption{Parameters used for data augmentation.}
\begin{tabular}{l|l|c}
\toprule
\textbf{Transformation}     & \textbf{Parameter}       & \textbf{Value}   \\ \midrule
Resize                      & Size                    & (64, 64)         \\ 
Random Flip Left-Right      & Probability             & Random           \\ 
Random Brightness           & Max Delta               & 0.2              \\ 
Random Contrast             & Range                  & (0.8, 1.2)       \\ 
Random Rotation             & Degrees                & 90°              \\ 
Sequence Length             & Frames                 & 20               \\ \bottomrule
\end{tabular}
\label{tab:Augment}
\end{table}

\subsection{Proposed Hybrid Model Architecture}

\begin{figure*}[!t]
    \centering
    \includegraphics[width=0.85\textwidth]{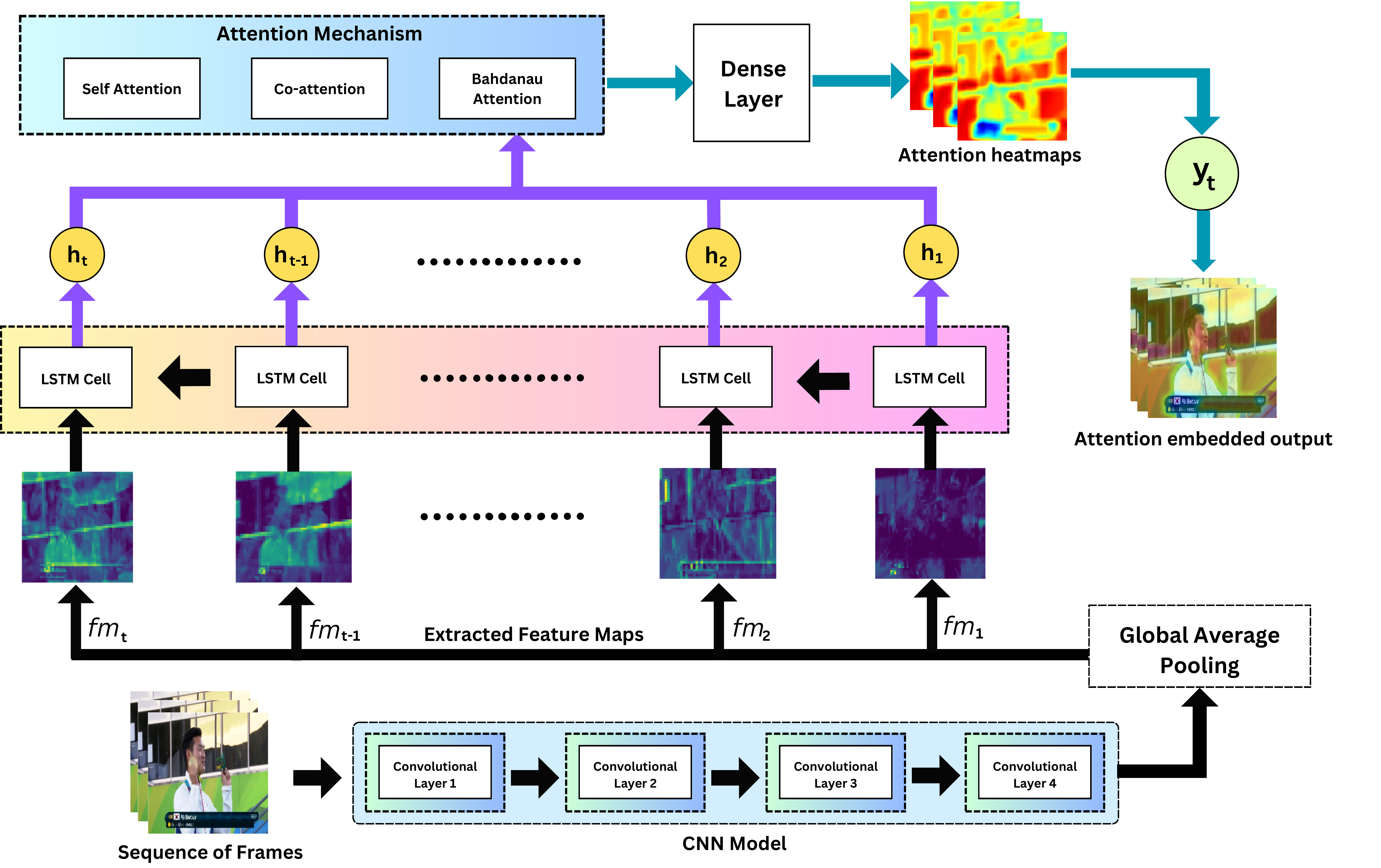}
    \caption{Proposed novel model architecture for video feature extraction and sequence processing. The illustration depicts a hybrid model that integrates CNN layers with max-pooling for feature extraction, followed by LSTM cells for sequence processing. Various attention mechanisms, including self-attention, co-attention, and Bahdanau attention, are incorporated to enhance temporal dependencies, culminating in a dense layer for output generation.}
    \label{model_arch}
\end{figure*}

In this study, we propose a novel hybrid model within the contrastive learning framework for video classification, specifically for MPAA rating classification. The proposed hybrid model leverages a combination of a CNN, LSTM, and an additional attention mechanism. The CNN is used for spatial feature extraction, the LSTM for temporal sequence modeling, and the attention mechanism for focusing on the most important parts of the sequence.
While the combination of CNN and LSTM, known as LRCN \cite{pandya2021segregating}, has been explored in existing studies, those models lack additional attention mechanisms. The fusion of these three architectures—CNN, LSTM, and attention—creates an efficient hybrid model well-suited for robust video classification. The overall architecture of the hybrid model is presented in Figure \ref{model_arch}.

\subsubsection{Convolutional Neural Network (CNN)}
The CNN part of the overall architecture extracts spatial features from the video frames, applying convolution and pooling layers:
\begin{equation}
    F^{(l)}(i, j) = \sigma \left( \sum_{m=1}^{M} \sum_{n=1}^{N} W_{mn}^{(l)} X(i + m, j + n) + b^{(l)} \right)
\end{equation}
where \( F^{(l)}(i, j) \) is the output of the \( l \)-th convolutional layer, \( W_{mn}^{(l)} \) are the filter weights, \( X(i + m, j + n) \) is the input frame, \( b^{(l)} \) is the bias term, and \( \sigma \) represents the ReLU activation function.
In our case, we employed the global average pooling on the extracted frames, which is defined as follows:
\begin{equation}
    F_{\text{GAP}} = \frac{1}{H \times W} \sum_{i=1}^{H} \sum_{j=1}^{W} F^{(l)}(i, j)
\end{equation}
where $H$ refers to the height and $W$ refers to the width of the extracted frames.

\subsubsection{Long Short-Term Memory (LSTM)}
The LSTM part processes the sequence of CNN-extracted feature vectors, modeling temporal dependencies of the video. The LSTM's equations are:
\begin{equation}
       \text{Input Gate:} \quad i_t = \sigma(W_i x_t + U_i h_{t-1} + b_i)
    \end{equation}
    \begin{equation}
       \text{Forget Gate: } f_t = \sigma(W_f x_t + U_f h_{t-1} + b_f)
    \end{equation}
    \begin{equation}
       \text{Cell State Update: } c_t = f_t \odot c_{t-1} + i_t \odot \tanh(W_c x_t + U_c h_{t-1} + b_c)
    \end{equation}
    \begin{equation}
        \text{Output Gate: } o_t = \sigma(W_o x_t + U_o h_{t-1} + b_o)
    \end{equation}    
    \begin{equation}
       \text{Hidden State: } h_t = o_t \odot \tanh(c_t)
\end{equation}
where \( x_t \) is the input feature vector at time step \( t \), \( h_t \) is the hidden state, \( c_t \) is the cell state, and \( (W, U, b) \) are learned weights and biases.

\subsubsection{Attention Mechanism}
In addition to the CNN and LSTM architectures, we incorporated an additional attention mechanism. In this study, we applied three distinct types of attention mechanisms: \emph{(a)} Self-attention \cite{vaswani2017attention}, \emph{(b)} Co-attention \cite{lu2016hierarchical}, and \emph{(c)} Bahdanau attention \cite{bahdanau2014neural}, tailored to the specific requirements of each contrastive learning method. The attention mechanism dynamically weights critical frames containing class indicators, enabling precise discrimination between MPAA classes by amplifying rating-defining spatiotemporal features.

\noindent\textbf{Self-attention Mechanism:} We leverage the self-attention mechanism for the Instance Discrimination contrastive learning framework. The self-attention mechanism allows the model to attend to different parts of a single input sequence to capture important temporal or spatial information. It computes attention scores $u_t$ within the same sequence to assign weights to relevant features. The attention scores for each time step in the sequence are computed using the dot product between a hidden state \( h_t \) and learned weights \( W \).
\begin{equation}      
u_t = \tanh(W h_t + b)
\end{equation} 
where \( W \) is the weight matrix for the attention mechanism, \( b \) is the bias term, and \( h_t \) is the hidden state at time step \( t \). The attention scores \( u_t \) are then passed through a softmax function to generate attention weights.       
\begin{equation}
        \alpha_t = \frac{\exp(u_t)}{\sum_{t'} \exp(u_{t'})}
\end{equation}
where \( \alpha_t \) is the attention weight for time step \( t \), and the denominator is the sum over all time steps in the sequence to normalize the attention weights. Finally, the context vector \( c \) is computed as a weighted sum of the hidden states, where the weights are given by the attention weights \( \alpha_t \).
    \begin{equation}
        c = \sum_{t} \alpha_t h_t
    \end{equation}
This context vector \( c \) captures the most relevant information from the sequence, which is then used for classification. 

\begin{figure}[!t]
    \centering
    \includegraphics[width=0.48\textwidth]{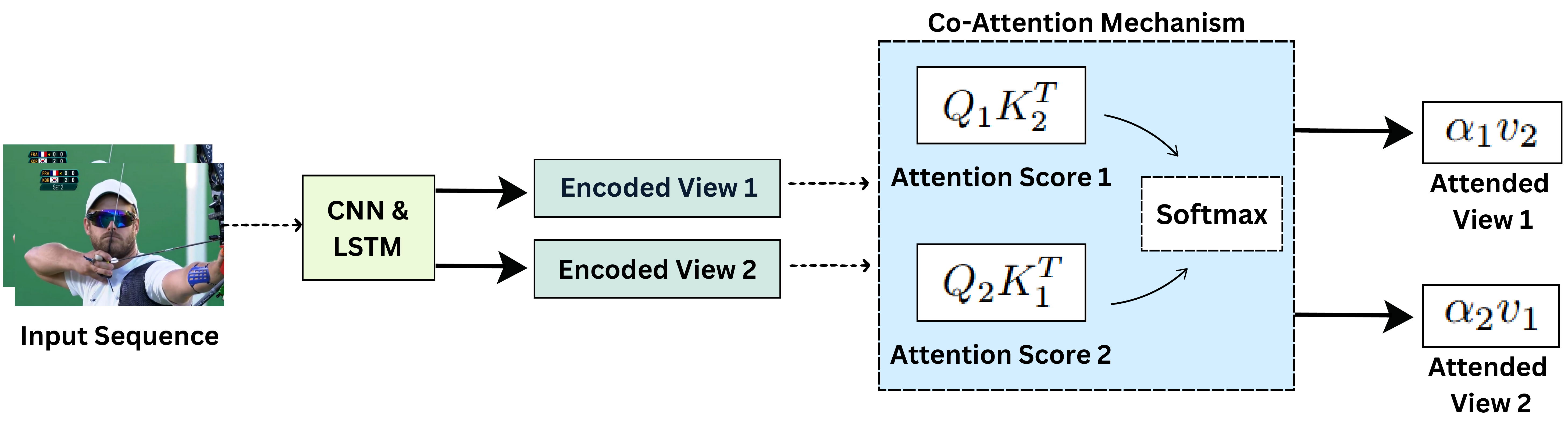}
    \caption{Illustration of the proposed co-attention mechanism. The input sequence is first processed by convolutional and recurrent (SoftmaxCNN and LSTM) layers to produce encoded representations from two views. Attention scores are computed for each view, which are then combined through the co-attention mechanism to generate attended representations that capture cross-view dependencies.}
    \label{fig:Co-attention}
\end{figure}

\noindent\textbf{Co-attention Mechanism:} The co-attention mechanism is employed for multi-view contrastive learning as depicted in Figure \ref{fig:Co-attention} . It computes attention scores between two different views (or inputs) and applies these scores to weight the feature representations from each view. This mechanism helps the model attend to relevant features by jointly learning attention across both views. Co-attention mechanism is described by the following components: \emph{Query}, \emph{Key}, and \emph{Value Representations}. For each view (e.g., view 1 and view 2), the query $(Q)$, key $(K)$, and value  $(V)$ representations are computed using learned weight matrices:
    \begin{equation}
    \begin{aligned}
        Q_1 &= W_Q v_1, \quad K_1 = W_K v_1, \quad V_1 = W_V v_1 \\
        Q_2 &= W_Q v_2, \quad K_2 = W_K v_2, \quad V_2 = W_V v_2
    \end{aligned}
    \end{equation}
where \( W_Q \), \( W_K \), and \( W_V \) are the weight matrices for the query, key, and value transformations, respectively, and
\( v_1 \) and \( v_2 \) are the input representations of the two views. The attention scores are computed by taking the dot product of the query of one view and the key of the other view.
   \begin{equation}
        \text{Attention Score 1} = Q_1 K_2^T
    \end{equation}
    \begin{equation}
        \text{Attention Score 2} = Q_2 K_1^T
    \end{equation}
The attention scores are then normalized using the softmax activation function to get the attention weights.
    \begin{equation}
        \alpha_1 = \text{Softmax}(\text{Attention Score 1})
    \end{equation}
    \begin{equation}
        \alpha_2 = \text{Softmax}(\text{Attention Score 2})
    \end{equation}
The attention weights ($\alpha_1$, $\alpha_2$) are used to weigh the value representations of the opposite views.
\begin{equation}
    \text{Attended View 1} = \alpha_1 v_2
\end{equation}
\begin{equation}
    \text{Attended View 2} = \alpha_2 v_1
\end{equation}
Thus, each view attends to relevant features from the other view, resulting in the attended representations for both views.

\noindent\textbf{Bahdanau Attention Mechanism:} The Bahdanau attention is used for the contextual contrastive learning method. This mechanism helps the model focus on the most important frames in the sequence. The attention scores are calculated for each frame in the sequence, and a context vector is computed as a weighted sum of the LSTM's hidden states. The attention mechanism is described by the following equations:
    \begin{equation}
        \text{Score Calculation:} \quad e_t = v^T \tanh(W_1 h_t + W_2 h_{Q})
    \end{equation}
where \( e_t \) is the unnormalized attention score for time step \( t \), \( h_t \) is the hidden state at time step \( t \), and \( h_{Q} \) is the query vector (in this case, the final hidden state of the LSTM). \( W_1\) and \(W_2\) are weight matrices, and \( v^T \) is the transpose of the view weight vector. Then, softmax is applied to normalize the attention scores into probabilities.
\begin{equation}
    \text{Softmax:} \quad \alpha_t = \frac{\exp(e_t)}{\sum_{t'} \exp(e_{t'})}
\end{equation}
Finally, the context vector \( c \) is computed as the weighted sum of the hidden states, where the weights \( \alpha_t \) reflect the importance of each hidden state.
\begin{equation}
    \text{Context Vector:} \quad c = \sum_{t} \alpha_t h_t
\end{equation}

\subsection{Contrastive Learning}
Contrastive learning \cite{chen2020simple} is an advanced machine learning technique used primarily in unsupervised \cite{xie2021detco} and semi-supervised \cite{chen2020big} settings to train models by teaching them to distinguish In the context of MPAA rating video classification, the contrastive learning approach is particularly effective for differentiating subtle distinctions between the four rating classes. Contrastive learning is uniquely suited for this task because it mitigates label scarcity by learning invariant representations from unlabeled or sparsely labeled data, while its ability to amplify fine-grained differences in content (e.g., brief violent sequences vs. sustained intensity) directly addresses the overlapping and subjective nature of MPAA rating criteria, enabling precise class separation. In this study, we used three different contrastive learning methods: \emph{(a)} Instance Discrimination \cite{chen2020simple, xie2021detco}, which differentiates individual data points; \emph{(b)} Multiview Contrastive Learning \cite{tian2020contrastive}, which harnesses multiple data representations; and \emph{(c)} Contextual Contrastive Learning \cite{sung2024contextrast}, focusing on relationships within data contexts. These methods collectively enhance our model's robustness and contextual awareness.

\subsubsection{Contrastive Pair Creation}
Contrastive pairs are generated to facilitate effective representation learning during model pre-training. These pairs consist of augmented variations of the same sequence, referred to as positive pairs. This approach enables the model to learn discriminative features by maximizing the similarity between positive pairs while distinguishing them from contrastive (negative) pairs. 

\noindent\textbf{Instance Discrimination:}
This contrastive learning approach generates two augmented views (frames) of the same video sequence for each input. The goal is to create positive pairs for instance discrimination, where both views come from the same instance. For each video sequence \( X \), two augmented sequences \( \tilde{X}_1 \) and \( \tilde{X}_2 \) are generated. This can be expressed as $(\tilde{X}_1, \tilde{X}_2)$,  where $\tilde{X}_1 = \text{Augment}(X)$ and $\tilde{X}_2 = \text{Augment}(X)$. 
 \( \text{Augment}(X) \) denotes the application of a series of transformations to the original video sequence \( X \). Hence, \( \tilde{X}_1 \) and \( \tilde{X}_2 \) are two different augmented versions of the same sequence, forming a positive pair for contrastive learning. 
     
\begin{figure}[!h]
    \centering
    \includegraphics[width=0.5\textwidth]{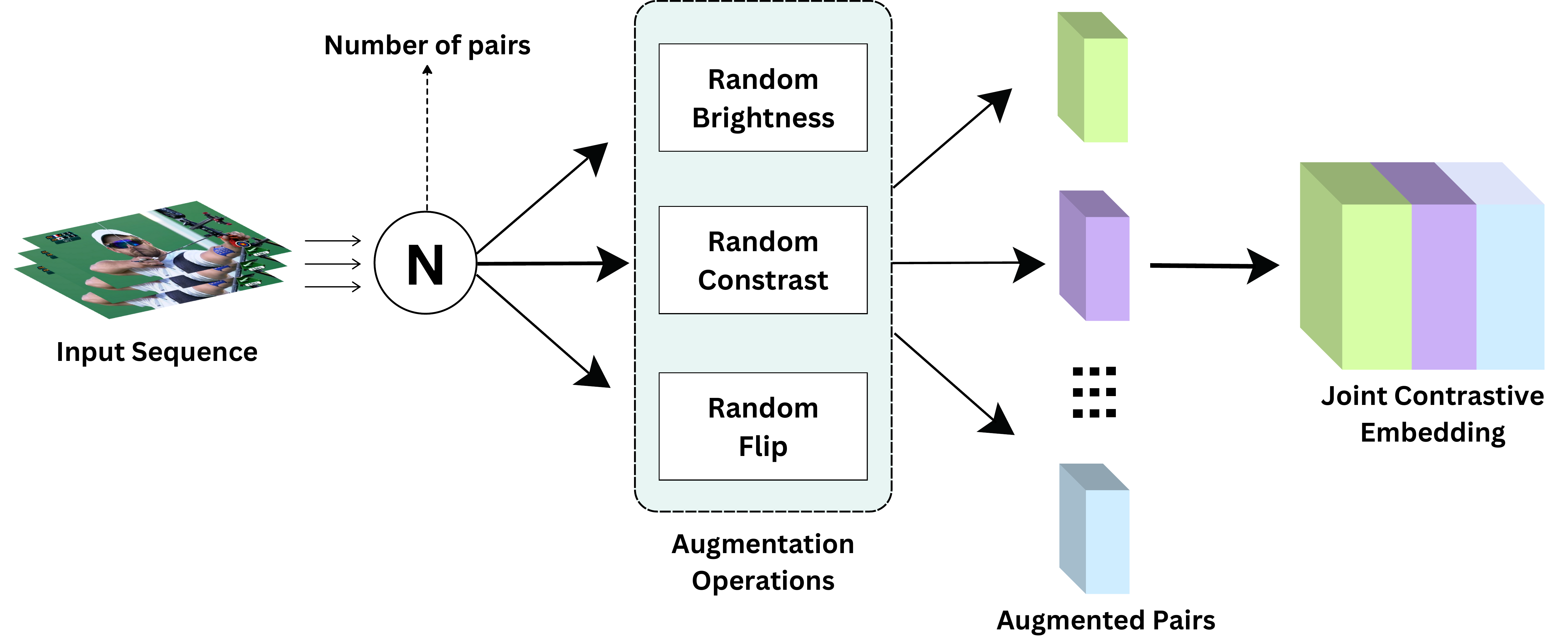}
    \caption{Generation of multi-view pairs in the contrastive learning framework. This process involves input sequence augmentation through techniques such as random flipping, brightness adjustment, and contrast modification. The augmented sequences are then integrated to enhance model robustness by increasing the diversity of the training data.}
    \label{fig:Multiview pair}
\end{figure}

\noindent\textbf{Multi-view Contrastive Learning:}
This framework creates multiple views of the same sequence. Each view is an augmented version of the input video sequence. This technique extends the contrastive learning paradigm by incorporating multiple positive views of the same instance. For each video sequence \( X \), a set of \( N \) augmented sequences \( \{\hat{X}_1, \hat{X}_2, \dots, \hat{X}_N\} \) are generated. 
\begin{equation}
\hat{X}_i = \text{Augment}(X) \quad \text{for} \quad i = 1, 2, \dots, N
\end{equation}
Here, \( N \) is the number of augmented views, and \( \hat{X}_i \) is the \( i \)-th augmented version of the video sequence \( X \). The objective is to pull the representations of all augmented views of the same video sequence together, as shown in Figure \ref{fig:Multiview pair}, while pushing away the representations of views from other sequences. 

\begin{figure}[!t]
    \centering
    \includegraphics[width=0.5\textwidth]{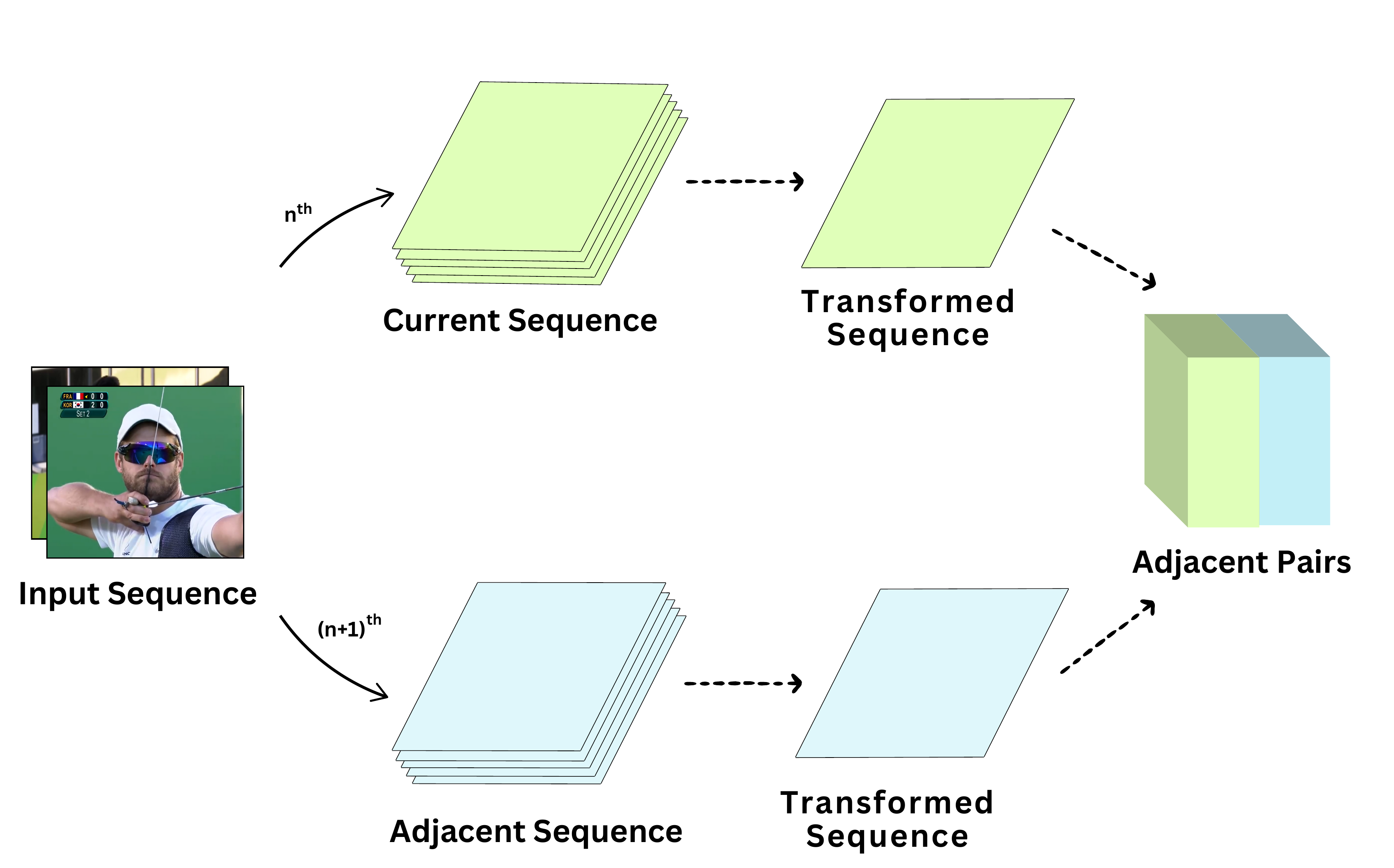}
    \caption{Contextual pair(Adjacent) creation in contextual contrastive learning. This figure illustrates the process of modifying input sequences through random brightness, contrast, and flipping adjustments, followed by the integration of augmented sequences. }
    \label{contextualpair}
\end{figure}

\noindent\textbf{Contextual Contrastive Learning:}
Contextual contrastive pairs are generated by including augmented versions of the same sequence (positive pairs) and adjacent sequences from the same video (contextual pairs), as depicted in Figure \ref{contextualpair}. The equations for constructing these pairs are as follows:
 \begin{equation}
    \begin{split}
        \textit{Augmented Pairs: } 
        \hat{X}_1 = \text{Augment}(X), \, \hat{X}_2 = \text{Augment}(X)
    \end{split}
\end{equation}
\begin{equation}
    \begin{split}
        \textit{Adjacent Pairs: }   (X_i, X_{i+1})
    \end{split}
\end{equation}
where $X_i$ and $X_{i+1}$ are from adjacent frames.
 
\subsubsection{Contrastive Loss}
In this study, we employed three different types of contrastive loss \cite{hadsell2006dimensionality} functions—NT-Xent, NT-Logistic, and Margin Triplet—to pre-train the models on unlabeled data. These loss functions enable the model to learn meaningful representations by maximizing the similarity between positive pairs while minimizing it for negative pairs, without requiring labeled supervision. 

\noindent\textbf{NT-Xent Loss:} The NT-Xent (Normalized Temperature-scaled Cross Entropy) loss function is used to encourage positive pairs to have higher similarity scores than negative pairs within a batch. The steps to compute the NT-Xent loss involve normalizing feature vectors, calculating the similarity matrix, and applying the temperature-scaled cross-entropy loss function.

To normalize feature vectors \( z_i \) and \( z_j \), we divide each vector by its norm to obtain unit-length representations:    
    \begin{equation}
        \hat{z}_i = \frac{z_i}{\|z_i\|}, \quad \hat{z}_j = \frac{z_j}{\|z_j\|}
    \end{equation}
    where \( \|z_i\| \) denotes the norm of vector \( z_i \), and \( \hat{z}_i \) and \( \hat{z}_j \) are the normalized feature vectors. After computing the normalized feature vectors, the similarity between the two normalized vectors \( \hat{z}_i \) and \( \hat{z}_j \) is computed as their dot product:   
    \begin{equation}
        S_{ij} = \hat{z}_i \cdot \hat{z}_j
    \end{equation}
where \( S_{ij} \) represents the similarity between vectors \( \hat{z}_i \) and \( \hat{z}_j \). Finally, the NT-Xent loss is calculated by applying a temperature-scaled cross-entropy function to the similarity scores for positive and negative pairs within a batch.
    \begin{equation}
        \mathcal{L}_{\text{NT-Xent}} = \frac{1}{2N} \sum_{i=1}^{2N} \left[ -\log \frac{\exp(S_{i,i+N} / \tau)}{\sum_{k \neq i} \exp(S_{i,k} / \tau)} \right]
    \end{equation}
where \( \tau \) represents a hyperparameter (temperature), \( i \) and \( i+N \) are the indices for positive pairs within the batch, and \( k \neq i \) represents indices for negative pairs in the batch. 
    

The temperature parameter \( \tau \) scales similarity scores, affecting the model's sensitivity to differences between positive and negative pairs. Lower \( \tau \) values enhance distinction, while higher values reduce sensitivity. The NT-Xent loss penalizes cases where positive pair similarity \( S_{i,i+N} \) is not significantly higher than negative pair similarities \( S_{i,k} \). Minimizing this loss encourages the model to maximize positive pair similarity while minimizing negative pair similarity.
 
\noindent\textbf{NT-Logistic Loss:} Similar to NT-Xent, the NT-Logistic loss also aims to separate the feature space into distinct clusters of similar and dissimilar pairs. However, it uses a logistic loss instead of the cross-entropy loss. The normalized feature vectors and similarity matrix are computed as described in NT-Xent loss. Finally, the NT-logistic loss is calculated by applying a logistic function to the similarity scores for positive pairs and negative pairs within a batch.
\begin{equation}
    \begin{split}
        \mathcal{L}_{\text{NT-logistic}} &= \frac{1}{2N} \sum_{i=1}^{2N} \Bigg[ \log\left(1 + \exp\left(-\frac{S_{i,i+N}}{\tau}\right)\right) \\
        &\quad + \sum_{k \neq i} \log\left(1 + \exp\left(\frac{S_{i,k}}{\tau}\right)\right) \Bigg]
    \end{split}
\end{equation}
where \( \tau \) is the temperature hyperparameter, \( i \) and \( i+N \) are the indices for positive pairs within the batch, and \( k \neq i \) represents indices for negative pairs in the batch. 

\begin{figure*}[!t]
    \centering
    \includegraphics[width=0.9\textwidth]{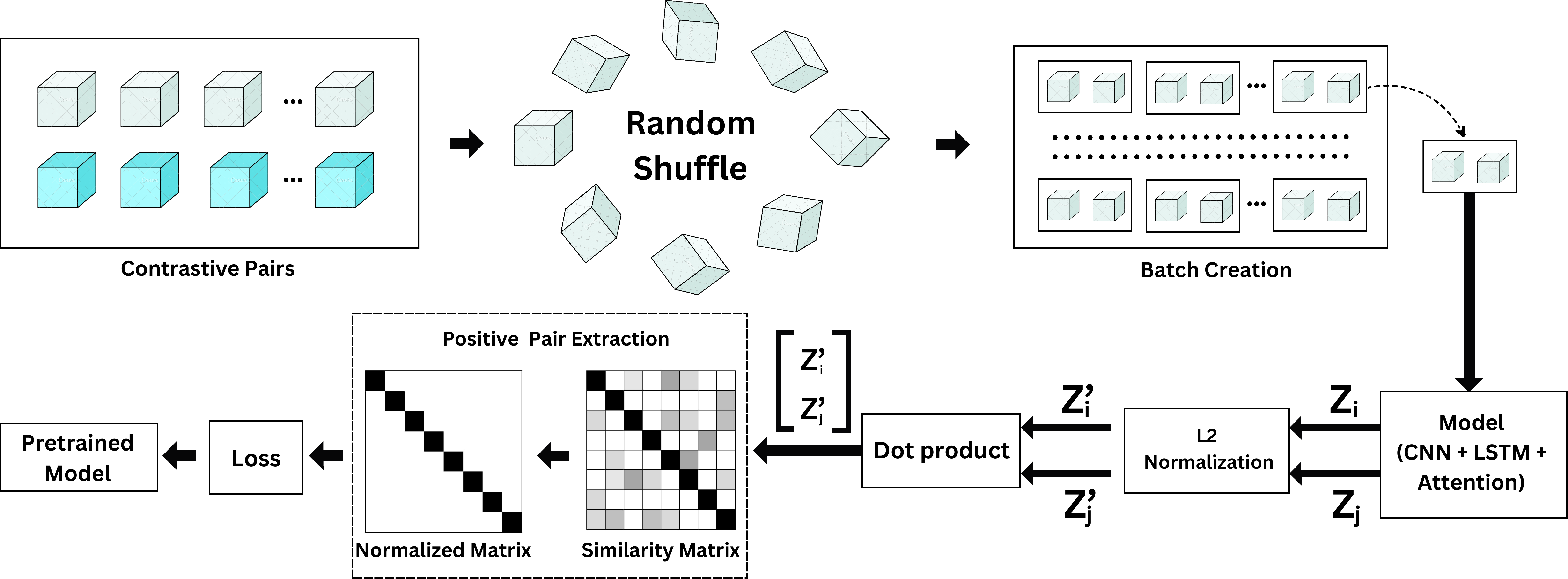}
    \caption{Pre-training Workflow in Contrastive Learning. This figure outlines the pre-training stages for contrastive learning, illustrating steps from the generation of contrastive pairs to the preparation of a pre-trained model. Key processes include pair shuffling, batch processing, normalization, and the computation of contrastive loss, leading to a model for fine-tuning.}
    \label{fig:Pretraining}
\end{figure*}

\noindent\textbf{Margin Triplet Loss:} The Margin Triplet loss encourages the model to learn embeddings such that the distance between an anchor and a positive sample (same class) is smaller than the distance between the anchor and a negative sample (different class) by at least a margin. At first, we normalize the feature vectors, which are then used to compute positive pair and negative pair distances. 
\begin{equation}
    \hat{a} = \frac{a}{\|a\|}, \quad \hat{p} = \frac{p}{\|p\|}, \quad \hat{n} = \frac{n}{\|n\|}
\end{equation}
\begin{equation} 
 d_{\text{positive}} = \|\hat{a} - \hat{p}\|^2, \quad d_{\text{negative}} = \|\hat{a} - \hat{n}\|^2
\end{equation}
where \(\hat{a}\), \(\hat{p}\) and \(\hat{n}\) represent the anchor vector, positive vector, and the negative vector, respectively. Hence, the triplet loss is computed as: 
\begin{equation}
    \mathcal{L}_{\text{triplet}} = \max(d_{\text{positive}} - d_{\text{negative}} + \text{margin}, 0)
\end{equation} 
where \emph{margin} is a hyperparameter that ensures that the negative pair distance is sufficiently larger than the positive pair distance by at least this value. Finally, the overall triplet loss is computed as the average over all triplets in the batch.
\begin{equation}
    \mathcal{L}_{\text{batch}} = \frac{1}{N} \sum_{i=1}^{N} \mathcal{L}_{\text{triplet}}^i
\end{equation}
where \( N \) is the batch size and \( \mathcal{L}_{\text{triplet}}^i \) is the loss for the \( i \)-th triplet.

A comparative analysis of model performance using different contrastive loss functions for MPAA rating video classification is discussed in Section \ref{result_section}.

\subsubsection{Model Training}
The overall training phase is divided into two parts: pre-training and fine-tuning. The pre-training process, illustrated in Figure \ref{fig:Pretraining}, leverages a contrastive loss function to effectively structure the feature space. During this phase, we extract $N$ pairs of embeddings $(\mathbf{z}_i, \mathbf{z}_j)$ and normalize them using L2 normalization to ensure their magnitudes are 1. These normalized embeddings are then concatenated to form a matrix $\mathbf{R} \in \mathbb{R}^{2N \times d}$. Next, we compute a pairwise cosine similarity matrix $\mathbf{S} \in \mathbb{R}^{2N \times 2N}$, where each entry represents the cosine similarity between embedding pairs. Positive pairs are identified along the diagonal of $\mathbf{S}$. To sharpen the similarity distribution and emphasize hard positives and negatives, the similarity scores are scaled using a temperature hyperparameter $\tau$. After pre-training, the fine-tuning phase adapts the model to specific tasks by further refining the accuracy and applicability of the learned representations. Unlike pre-training, contrastive loss is not used during this phase. Instead, we employ categorical cross-entropy as the primary loss function to quantify the error between predicted probabilities and actual class labels, thereby enhancing classification performance.

\section{Experiments}

\subsection{Setup}

\noindent \textbf{Implementation Details:}
Our proposed contrastive learning framework is built on an LRCN architecture, integrating CNNs for spatial feature extraction, LSTMs for temporal sequence modeling, and an attention mechanism to dynamically prioritize significant frames.
To enhance feature learning, we employ three distinct contrastive loss functions: NT-Xent, NT-Logistic, and Margin Triplet. The model undergoes pre-training for 20 epochs with a batch size of 32. We employ specialized pre-training techniques using gradient clipping to prevent LSTM units from overtraining, thereby enhancing the stability and convergence of our model. During the fine-tuning phase, we optimize the model using cross-entropy loss and the Adam optimizer with a learning rate of 1e-5. Due to computational constraints, the batch size is set to 4 for fine-tuning. Additionally, we implement early stopping with a patience of 10 epochs to prevent overfitting.
All model training is conducted on a single Google Cloud v2-8 TPU, ensuring efficient computation The implementation is carried out using the \emph{PyTorch} framework for seamless experimentation and model training.

\begin{table*}[!t]
\centering

\caption{Performance Summary of Various Contrastive Learning Methods
Methods apply different backbone architectures and loss functions across multiple evaluation metrics}
\label{performance2}
\resizebox{0.95\textwidth}{!}{
\begin{tabular}{l l l c c c c c}
\toprule
\textbf{Contrastive Method} & \textbf{Backbone} & \textbf{Loss} & \textbf{Accuracy (\%)} & \textbf{Precision (\%)} & \textbf{Recall (\%)} & \textbf{F1 (\%)} & \textbf{AUC}  \\ \midrule
\multirow{9}{*}{\textbf{Instance Discrimination}} & ResNet3D-50 & NT-Xent & 61.00 & 60.00 & 59.00 & 59.00 & 0.80 \\
& ResNet3D-50 & NT-Logistic & 54.67 & 57.00 & 53.00 & 52.00 & 0.77 \\
& ResNet3D-50 & Margin Triplet & 56.00 & 54.75 & 55.00 & 54.50 & 0.80 \\
& LRCN & NT-Xent & 76.00 & 77.00 & 77.00 & 75.00 & 0.92 \\
& LRCN & NT-Logistic & 71.00 & 71.00 & 72.00 & 69.00 & 0.90 \\
& LRCN & Margin Triplet & 65.33 & 64.00 & 62.00 & 63.00 & 0.87 \\
& LRCN + Attention & NT-Xent & \textbf{77.33} & \textbf{78.65} & \textbf{77.33} & \textbf{77.37} & \textbf{0.93} \\
& LRCN + Attention & NT-Logistic & 69.33 & 70.41 & 69.33 & 69.17 & 0.90 \\
& LRCN + Attention & Margin Triplet & 72.00 & 71.60 & 72.00 & 71.57 & 0.93 \\ \midrule
\multirow{9}{*}{\textbf{Multi-View}} & ResNet3D-50 & NT-Xent & 48.00 & 50.00 & 49.00 & 42.00 & 0.69 \\
& ResNet3D-50 & NT-Logistic & 53.33 & 50.88 & 53.33 & 52.00 & 0.76 \\
& ResNet3D-50 & Margin Triplet & 53.00 & 51.00 & 53.33 & 51.62 & 0.76 \\
& LRCN & NT-Xent & 43.00 & 32.61 & 42.67 & 36.50 & 0.69 \\
& LRCN & NT-Logistic & 47.00 & 37.36 & 46.67 & 41.49 & 0.69 \\
& LRCN & Margin Triplet & 43.00 & 32.61 & 43.00 & 37.00 & 0.69 \\
& LRCN + Attention & NT-Xent & 56.00 & 50.00 & 56.00 & 53.00 & 0.72 \\
& LRCN + Attention & NT-Logistic &\textbf{76.00} & \textbf{81.32} &\textbf{76.00} & \textbf{78.57} & \textbf{0.93} \\
& LRCN + Attention & Margin Triplet & 55.00 & 60.70 & 54.64 & 57.54 & 0.75 \\ \midrule
\multirow{9}{*}{\textbf{Contextual}} & ResNet3D-50 & NT-Xent & 72.00 & 72.08 & 71.28 & 74.46 & 0.93 \\
& ResNet3D-50 & NT-Logistic & 67.00 & 72.00 & 67.00 & 67.00 & 0.93 \\
& ResNet3D-50 & Margin Triplet & 78.57 & 80.13 & 78.57 & 79.34 & 0.95 \\
& LRCN & NT-Xent & 80.00 & 81.45 & 80.00 & 80.30 & 0.96 \\
& LRCN & NT-Logistic & 81.16 & 82.56 & 81.17 & 81.42 & 0.96 \\
& LRCN & Margin Triplet & 82.47 & 83.51 & 82.47 & 83.00 & 0.97 \\
& LRCN + Attention & NT-Xent & 82.00 & 84.13 & 79.22 & 82.00 & 0.96 \\
& LRCN + Attention & NT-Logistic & \textbf{88.00} &\textbf{89.33} & \textbf{87.01} & \textbf{88.15} & \textbf{0.98} \\
& LRCN + Attention & Margin Triplet & 85.71 & 87.33 & 85.06 & 86.18 & 0.96 \\
\bottomrule
\end{tabular}
}
\end{table*}

\begin{figure*}[t!]
    \centering
    \resizebox{0.95\textwidth}{!}{
    \begin{subfigure}[b]{0.3\linewidth}
        \centering
        \includegraphics[width=\textwidth]{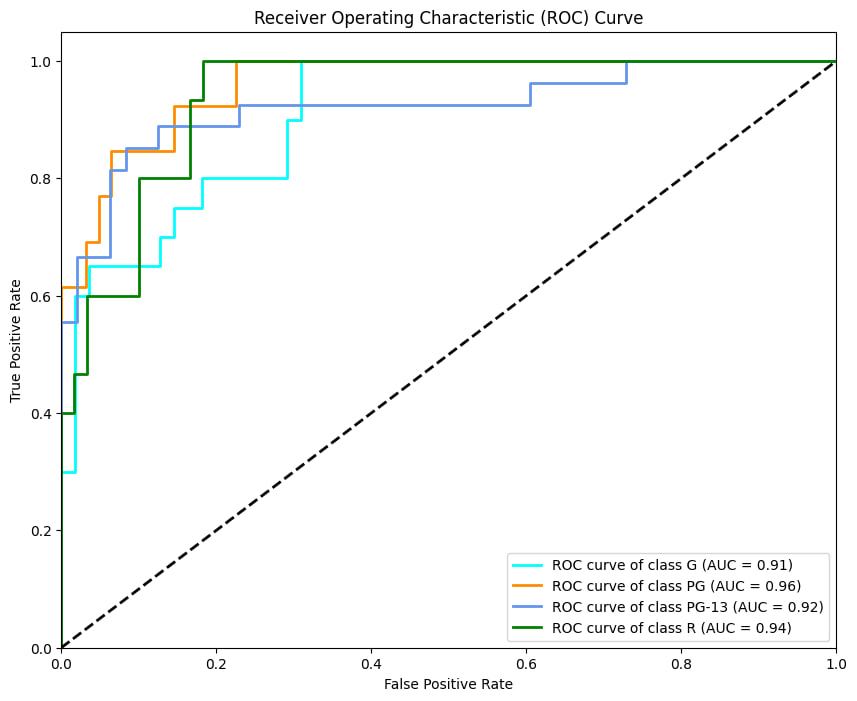}
        \caption{Instance Discrimination}
        \label{fig:Conf_Swin}
    \end{subfigure}
    \hfill
    \begin{subfigure}[b]{0.3\linewidth}
        \centering
        \includegraphics[width=\textwidth]{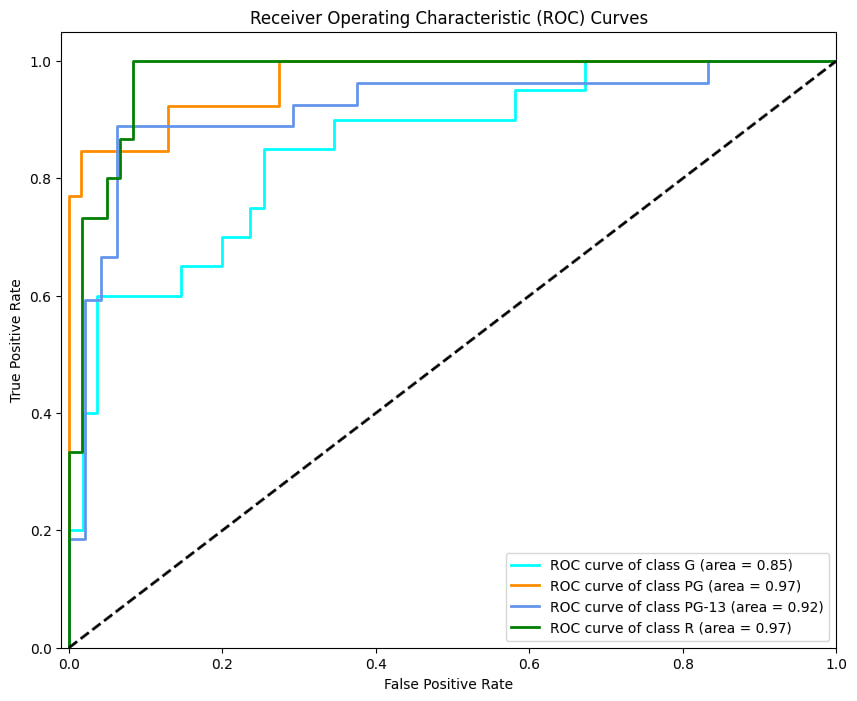}
        \caption{Multi-view Contrastive Learning}
        \label{fig:Conf_vit1}
    \end{subfigure}
    \hfill
    \begin{subfigure}[b]{0.3\linewidth}
        \centering
        \includegraphics[width=\textwidth]{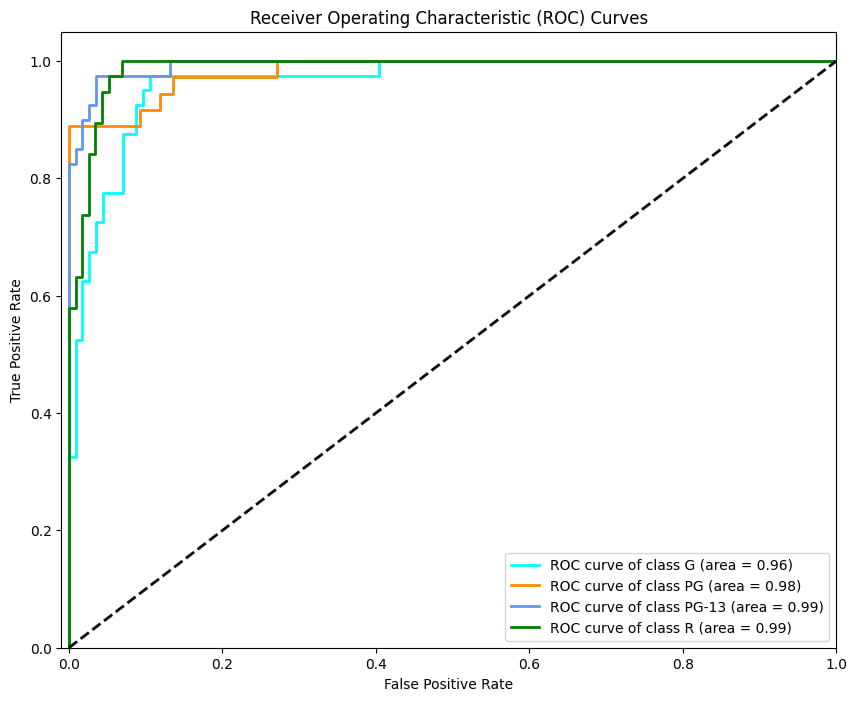}
        \caption{Contextual Contrastive Learning}
        \label{fig:Conf_Sota}
    \end{subfigure}
   }
    \caption{Receiver Operating Characteristic (ROC) curves for the best models from three distinct contrastive learning methods using our proposed hybrid architecture (LRCN + Attention).}
    \label{roc_curves}
\end{figure*}


\begin{table}[t!]
\centering
\caption{Comparison of the number of parameters, execution time, and classification accuracy across different backbones for three distinct contrastive learning methods. Arrows ($\uparrow$ for higher, $\downarrow$ for lower) indicate preferable scores.}
\label{tab:comparison}
\resizebox{0.5\textwidth}{!}{
\renewcommand{\arraystretch}{1.2}
\begin{tabular}{c|l|c|c|c}
\toprule
\textbf{Method} & \textbf{Backbone} & \textbf{Time $\downarrow$} & \textbf{Param $\downarrow$} & \textbf{Acc(\%) $\uparrow$} \\ 
\midrule
 \multirow{2}{*}{Instance} & ResNet3D-50  & 24.04 s & 14.38 M & 61.00 \\
 \multirow{2}{*}{Discrimination} & LRCN & 53.23 s & \textbf{5.08 M} & 76.00 \\
 & \cellcolor{gray!20}LRCN + self-Attention & \cellcolor{gray!20}\textbf{20.81 s} & \cellcolor{gray!20}5.14 M & \cellcolor{gray!20}\textbf{77.33} \\
\midrule
 & ResNet3D-50 & \textbf{78.75 s} & 14.38 M & 53.33 \\

Multi-view & LRCN &  261.43 s & 8.93 M & 47.00 \\

 & \cellcolor{gray!20}LRCN + Co-attention & \cellcolor{gray!20}255.63 s & \cellcolor{gray!20}\textbf{8.91 M} & \cellcolor{gray!20}\textbf{76.00} \\
\midrule
 & ResNet3D-50 & 52.00 s & 14.91 M & 78.57 \\

Contextual & LRCN & 8.10 s & 0.50 M & 82.47 \\

 & \cellcolor{gray!20}LRCN + Bahdanau attention &  \cellcolor{gray!20}\textbf{7.10 s} & \cellcolor{gray!20}\textbf{0.50 M} & \cellcolor{gray!20}\textbf{88.00} \\
\bottomrule
\end{tabular}
}
\end{table}

\noindent \textbf{Evaluation:}
To comprehensively assess the performance of our proposed model alongside existing architectures, we conducted evaluations on a 25\% unseen test set. The assessment utilized standard classification metrics, including accuracy, precision, recall, F1 score, and AUC, to measure both effectiveness and discriminative capability.
Beyond classification performance, we also compared models in terms of execution time and total trainable parameters, providing a holistic analysis of computational efficiency. These evaluations highlighted the advantages of integrating a lighter backbone and an attention mechanism, demonstrating improved efficiency without compromising accuracy. Additionally, our findings offer valuable insights into optimizing model design for enhanced operational and computational performance.

\subsection{Results and Analysis}
\label{result_section}
This section provides a comprehensive analysis of the experimental results for MPAA rating classification from video clips. We employed three distinct contrastive learning methods: Instance Discrimination, Multi-View Contrastive Learning, and Contextual Contrastive Learning. Table \ref{performance2} presents a detailed performance comparison of our proposed model architecture against existing models for video classification. The evaluation is conducted across five key metrics: accuracy (Acc.), precision (Pre.), recall (Rec.), F1 score (F1), and AUC score (AUC), offering a thorough assessment of model effectiveness.

\begin{table}[!t]
\centering
\caption{Comparison of our proposed methodology (LRCN + Attention) with existing methods for video classification on the MPAA classification task.}
\label{comp}
\resizebox{0.5\textwidth}{!}{
\renewcommand{\arraystretch}{1.2}
\begin{tabular}{l l c c c}
\toprule
\textbf{Model} & \textbf{Dataset} & \textbf{\# Class} & \textbf{Param} & \textbf{Acc(\%)} \\ 
\midrule
    InceptionV3 \cite{9089897} & Violent Scenes Dataset & 3 & 23.9 M & 76.1 \\
    ResNet3D-50 \cite{kolarik2023detecting} & Our Dataset & 4 & 14.38 M & 78.6 \\  
    LRCN \cite{uzzaman2022lrcn} & Our Dataset & 4 & 0.5 M & 82.5 \\ 
    \rowcolor{gray!20} LRCN + Attention & Our Dataset & 4 & \textbf{0.5 M} & \textbf{88.0} \\ 
\bottomrule
\end{tabular}
}
\end{table}

The key observations from Table \ref{performance2} are as follows: 
\textbf{\emph{(1)}} Contextual contrastive learning consistently outperforms other contrastive learning methods across different backbone architectures, demonstrating its ability to effectively leverage contextual data relationships for improved classification.
 \textbf{\emph{(2)}} For instance discrimination, the NT-Xent loss function has been shown to yield superior performance. In multi-view contrastive learning, the NT-logistic loss demonstrates better classification results. Meanwhile, in contextual contrastive learning, the margin triplet loss has been found to be the most effective overall. These findings suggest that the choice of loss function should be carefully aligned with the underlying contrastive learning objective to maximize performance.
\textbf{\emph{(3)}} Our proposed model architecture, which integrates the LRCN (LSTM + CNN) backbone with an attention mechanism, consistently outperforms both ResNet3D-50 and LRCN across all three contrastive learning frameworks, demonstrating its superior ability to extract and learn critical features. Moreover, it maintains stable performance across all evaluation metrics—accuracy, precision, recall, F1 score, and AUC—highlighting its robustness and effectiveness compared to existing video classification models.
\textbf{\emph{(4)}} Overall, our proposed approach (LRCN + attention), when combined with a contextual contrastive learning framework using NT-Logistic loss, achieves state-of-the-art classification scores—88\% accuracy and an F1 score of 88.15\%—establishing a new benchmark for MPAA rating classification from video clips.
Figure \ref{roc_curves} presents the Receiver Operating Characteristic (ROC) curves for the proposed LRCN + Attention architecture, evaluated across three different contrastive learning methods and three distinct loss functions.

\begin{figure*}[!t]
    \centering
    \includegraphics[width=0.8\textwidth]{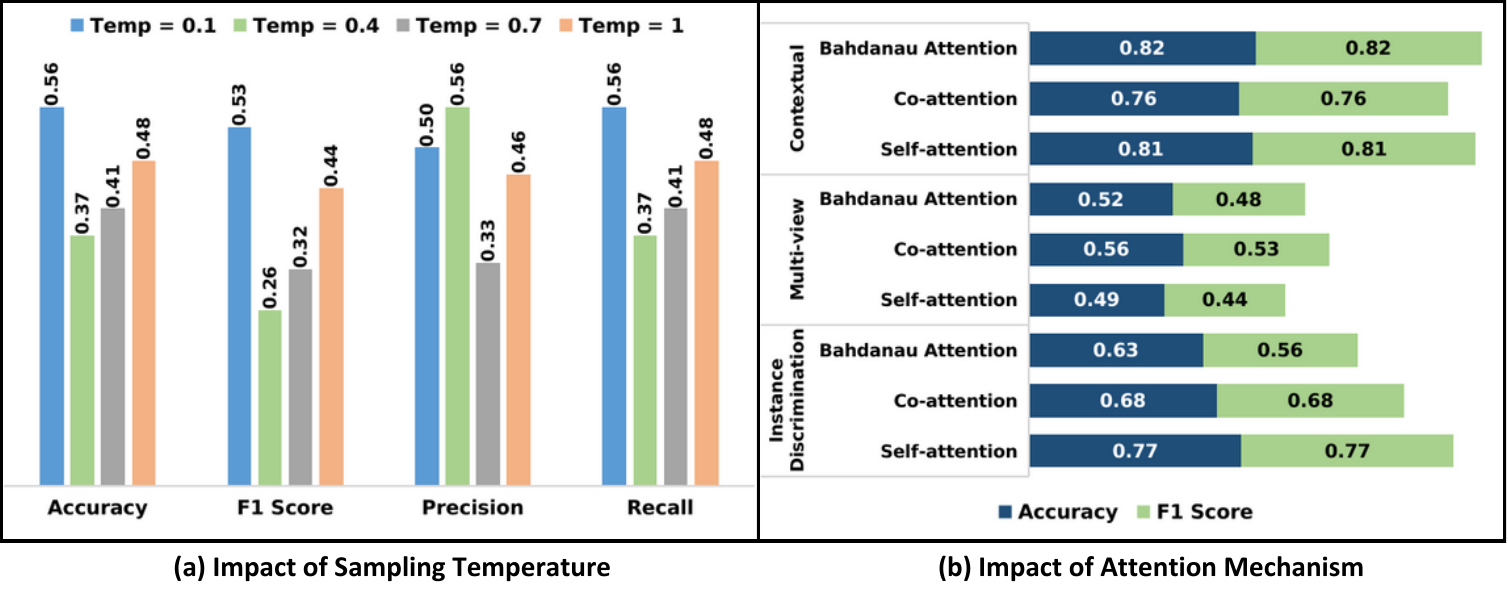}
    \caption{Illustration representing the impact of \textbf{(a)} varying sampling temperature of the loss function and \textbf{(b)} varying attention mechanism for the proposed model architecture.}
    \label{fig:ablation}
\end{figure*}

Table \ref{tab:comparison} compares different model architectures across various contrastive learning methods, evaluating each model based on execution time, the number of trainable parameters, and classification accuracy. The results indicate that the proposed LRCN + attention methodology requires less execution time and fewer trainable parameters while achieving significantly higher accuracy. For instance, within the contextual contrastive learning framework, combining LRCN with Bahdanau attention yields the highest accuracy of 88\%, outperforming other backbone architectures with a lower execution time (7.10 s) and fewer parameters (0.5 million). This demonstrates its strong performance and efficiency.

Finally, we present a comparison between our proposed model architecture and other notable architectures for MPAA rating classification from video scenes in Table \ref{comp}. As observed in the table, the proposed LRCN + attention method achieves the highest accuracy of 88\% while utilizing only 0.5 million parameters, surpassing existing approaches and models in both performance and efficiency.

\subsection{Ablation Studies}

To evaluate the impact of individual components in our experiments, we conduct ablation studies focusing on two key aspects: (a) sampling temperature of the loss function and (b) the type of attention mechanism used.

\noindent \textbf{Impact of Sampling Temperature:}
We first investigate how varying the sampling temperature in the loss function affects the overall classification performance. Specifically, we present the comparison of temperature values for the NT-Xent loss. We experimented with four distinct temperatures: 0.1, 0.4, 0.7 and 1.0, as presented in Figure \ref{fig:ablation}(a). We observe that a temperature setting of 0.1 yields the highest accuracy (56.00\%). Increasing the temperature to 0.4 drastically reduces the accuracy. Temperatures of 0.7 and 1.0 show moderate improvements over 0.4 but do not surpass the performance at 0.1, underscoring that a low temperature strikes an optimal balance between capturing informative negatives and maintaining robust classification metrics. Therefore, we use a sampling temperature of 0.1 for all our experiments.

\noindent \textbf{Impact of Attention Mechanisms:}
Beyond sampling temperature, we also investigate the impact of different attention mechanisms—self-attention, co-attention, and Bahdanau attention—on video classification performance. We evaluate model performance using various contrastive learning methods across all three attention mechanisms. For a fair ablation comparison, we employ the NT-Xent loss function in all evaluation settings. As shown in Figure \ref{fig:ablation}(b), the Bahdanau attention mechanism achieves the highest performance within the contextual contrastive learning framework. The co-attention mechanism records the highest accuracy and F1 score, highlighting its effectiveness in aligning and integrating multi-view features. Meanwhile, the self-attention mechanism excels in instance discrimination tasks, demonstrating its strong capability in this domain. Based on these findings, we select each attention mechanism according to the specific requirements of the contrastive learning approach to optimize performance.

\subsection{Web Implementation}
\begin{figure}[!t]
    \centering
    \includegraphics[width=0.49\textwidth]{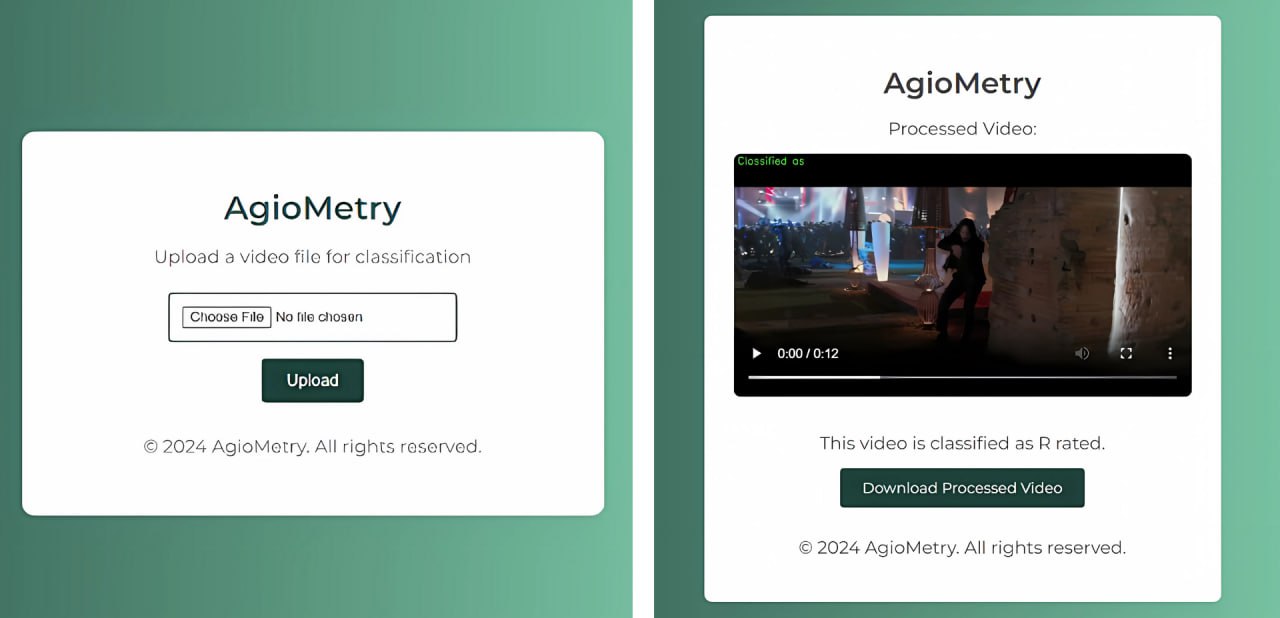}
    \caption{Interface of the video classification web application, which employs the trained LRCN + Attention model for MPAA rating classification of video content.}
    \label{fig:website}
\end{figure}

Finally, we deployed our best-performing model, LRCN + Attention, as a web application, showcasing its practical effectiveness in real-world deployment. The application is developed using Flask API, a Python-based web framework, ensuring a seamless and efficient user experience. Users can upload a video clip, which is then processed and classified into one of the MPAA rating categories (G, PG, PG-13, and R). Figure \ref{fig:website} illustrates the web interface of the developed application, highlighting its user-friendly design and functionality.


\section{Discussion}

The Motion Picture Association of America (MPAA) rating system, which classifies films into categories such as G (General Audiences), PG (Parental Guidance), PG-13 (Parents Strongly Cautioned), and R (Restricted), plays a crucial role in aiding families to make informed viewing choices. However, existing video classification techniques and datasets are not well-suited to accurately predict these MPAA ratings for video content. To address this gap, we first curated a specialized dataset tailored to this specific problem. We then proposed a novel hybrid model architecture within a contrastive learning framework that integrates an LRCN backbone with an attention mechanism to enhance learning efficacy. For different types of contrastive learning, we implemented specific attention types: self-attention, for instance, discrimination; co-attention for multi-view, and Bahdanau attention for contextual contrastive learning. Additionally, we explored three distinct contrastive loss functions—NT-Xent, NT-Logistic, and Margin Triplet—to optimize the embedding quality during training, ensuring model stability and convergence with the aid of gradient clipping.

Our proposed LRCN + attention architecture consistently outperforms existing video classification techniques across all types of contrastive learning. Notably, the contextual contrastive learning configuration utilizing an LRCN backbone combined with a Bahdanau attention mechanism and NT-Logistic loss has demonstrated the most superior performance among all evaluated models. Moreover, this model configuration not only delivers enhanced accuracy but also achieves this with fewer parameters and reduced execution time compared to existing models, highlighting its efficiency and effectiveness in handling complex video classification tasks in real-life scenarios.

\noindent \textbf{Limitations:}
Despite the significant advancements achieved, certain limitations remain. One major limitation is the exclusion of advanced contrastive learning approaches, such as Hierarchical and Spatiotemporal contrastive learning, which could potentially improve video classification efficiency. However, due to resource constraints, we were unable to explore these methodologies. Additionally, our proposed approach has not been evaluated on other video classification tasks, such as event detection or facial expression recognition. Furthermore, conducting comprehensive experiments on large-scale datasets like UCF50 \cite{reddy2013recognizing} and UCF101 \cite{soomro2012ucf101} was not feasible due to computational limitations. Moreover, throughout our experiments, we did not incorporate speech modalities, which could further enhance classification performance.

\section{Conclusion}
This study introduces a novel and practical approach for automated MPAA rating classification, addressing the growing need for scalable, efficient, and accurate content filtering in today’s digital media landscape. By leveraging a hybrid architecture within a contrastive learning framework, our model effectively integrates CNN, LSTM, and an attention mechanism to capture both spatial and temporal features for nuanced video classification. Additionally, our approach incorporates contrastive learning techniques—instance discrimination, multi-view, and contextual contrastive learning—alongside three distinct loss functions: NT-Xent, NT-Logistic, and Margin Triplet, leading to well-separated, robust embeddings for reliable classification. The lightweight design of our model, characterized by its compact size and reduced inference time, makes it highly suitable for real-time deployment, even on devices with limited computational resources.

Through extensive experimentation, we found that our LRCN + Attention architecture, when applied within a contextual contrastive learning framework, delivered the most accurate classification results, particularly in distinguishing between closely related rating categories such as PG-13 and R. Ultimately, our work bridges the gap between high-performance classification and practical usability by integrating novel architectural components and advanced contrastive learning strategies. By emphasizing both efficiency and real-time functionality, our approach sets a new benchmark for automated MPAA video classification, contributing to a safer and more responsible digital media environment.

\noindent \textbf{Future Works:}
Future research can explore additional contrastive learning frameworks, such as hierarchical and spatiotemporal contrastive learning, to further enhance video classification performance. Additionally, integrating large pre-trained models, such as ResNet-101 \cite{singh2022detection} and ResNet-152 \cite{mihanpour2020human} as feature extractors could improve spatial representation capabilities. Investigating transformer-based architectures for better temporal modeling and more advanced attention mechanisms may also lead to further improvements. These advancements would contribute significantly to the field of video classification, particularly in MPAA rating classification.

\bibliographystyle{IEEEtran}
\bibliography{example}

\end{document}